\documentclass[11pt]{article}

\usepackage[final]{acl}

\usepackage{times}
\usepackage{latexsym}

\usepackage[T1]{fontenc}
\usepackage{textcomp}
\usepackage[utf8]{inputenc}

\usepackage{microtype}

\usepackage{inconsolata}

\usepackage{graphicx}

\usepackage{amsmath}
\usepackage{stfloats}
\usepackage{booktabs}
\usepackage{tikz}
\usetikzlibrary{positioning, arrows.meta}
\usepackage{adjustbox}
\usepackage{hyperref}
\usepackage{booktabs}
\usepackage{tabularx}
\usepackage{array}

\title{The Linguistic Structure of Dependency-Conditioned Embeddings}

\author{
  \textbf{Bach Phan-Tat}\textsuperscript{1},
  \textbf{Kris Heylen}\textsuperscript{1,2},
  \textbf{Dirk Geeraerts}\textsuperscript{1},\\
  \textbf{Stefano De Pascale}\textsuperscript{1,3},
  \textbf{Tung Kieu}\textsuperscript{4},
  \textbf{Dirk Speelman}\textsuperscript{1}
  \\[1ex]
  \textsuperscript{1}Department of Linguistics, KU Leuven\\
  \textsuperscript{2}Instituut voor de Nederlandse Taal\\
  \textsuperscript{3}Department of Linguistics, UC Louvain\\
  \textsuperscript{4}Department of Computer Science, Aalborg University\\[1ex]
  \small \textbf{Correspondence:} \href{mailto:ttbach.phan@kuleuven.be}{ttbach.phan@kuleuven.be}
}

\begin{document}
\maketitle
\begin{abstract}
Dependency-based distributional models capture syntactically informed lexical similarities, but the geometric organisation of dependency-conditioned lexical representations remains underexplored. We investigate the global and local geometry of lemma–dependency relation embeddings using 4 representations trained on WikiText-103: 2 count-based PPMI configurations and 2 Skip-Gram with Negative Sampling (SGNS) configurations. At the global level, we analyse dependency-relation centroids, control for lexical identity through within-lemma centring, and examine their alignment with part-of-speech (POS) distributions. At the local level, we evaluate cross-relation geometric consistency, cross-method agreement, lexical similarity, and polysemous usages. Results show that global geometry is structured strongly by the context POS across all representations. Locally, broader lexical similarity structures are partially preserved across dependency relations, while immediate neighbourhoods vary substantially, highlighting different usages. PPMI-based representations exhibit greater cross-relation consistency than SGNS-based representations, with cross-method agreement varying by relation. These findings demonstrate that dependency-conditioned embeddings exhibit systematic global linguistic organisation alongside relation-sensitive local lexical structures, with the extent of geometric consistency varying across representation methods.

\end{abstract}

\section{Introduction and Related Work}
Distributional semantic models (DSMs) represent lexical meaning through patterns of word co-occurrence. The captured semantic information depends not only on the learning algorithm but also on the definition of the linguistic context. In particular, whereas linear-context embeddings exhibit stronger topical associations, dependency-based models, which use grammatical relations rather than linear proximity to define contexts, produce representations that emphasise functional similarities between words \cite{levy_dependency-based_2014, pado_dependency-based_2007, baroni_distributional_2010}. Beyond the broad distinction between linear and syntactic contexts, individual dependency relations may also provide complementary information about lexical meaning \cite{grefenstette_conquering_2007}. \citet{mouton_unsupervised_2009} constructed multiple semantic spaces from distinct syntactic relations and provided preliminary observations suggesting that the nearest neighbours of a polysemous word can vary across these spaces, highlighting different usages. 

Previous research has explored dependency conditioned lexical meaning through several representational approaches. One approach constructs separate semantic spaces for individual syntactic relations \cite{mouton_unsupervised_2009} using count-based methods, enabling relation-specific analyses but without a common coordinate system for directly examining cross-relation geometry. Through their Distributional Memory framework, \citet{baroni_distributional_2010} introduced a shared \textbf{word--link by word} semantic space in which lexical items conditioned on different dependency relations are jointly represented. They provided an initial demonstration of its linguistic utility through verb classification based on inter-slot similarities, leaving the broader geometric organisation of the shared space largely unexplored. Other approaches have incorporated dependency relations into contextual features to learn lemma-level embeddings \cite{levy_dependency-based_2014}.

While these studies demonstrate the usefulness of syntactically-structured representations, their primary focus has been on semantic applications rather than systematically characterising the geometric organisation of dependency-conditioned lexical units. In particular, relatively little attention has been given to how dependency relations are organised within a shared semantic space and how lexical similarity structures are preserved or differentiated across relations. An additional question concerns the extent to which this geometric organisation depends on the representation method. Count-based and predictive distributional models have been extensively compared in terms of semantic performance and representation properties, including models employing dependency contexts \cite{baroni_dont_2014, gamallo_comparing_2017, levy_improving_2015, riedl_theres_2017, lenci_comparative_2022}. However, these studies have primarily focused on lemma-level representations rather than dependency-conditioned lexical units themselves. The present study therefore investigates dependency-conditioned lexical representations at two complementary geometric levels: the global organisation of dependency relations within a shared semantic space and the local organisation of lexical similarities within and across relations. By comparing count-based and predictive representations, we further examine the extent to which these structures are consistent across embedding methods. The findings aim to inform the interpretation and selection of dependency-conditioned representations for lexical semantic research.

We investigate 4 representations derived from dependency-based co-occurrence information:
\begin{itemize}
    \item count-based representations constructed using 5000 most frequent context features; 
    \item count-based representations constructed using all context features but reduced to 300 dimensions through Singular Value Decomposition (SVD); 
    \item target embeddings obtained from a reversed dependency SGNS configuration, in which \texttt{lemma-relation} pairs serve as prediction targets;
    \item context embeddings obtained from original dependency SGNS.
\end{itemize}

All 4 represent \texttt{lemma-relation} pairs within a shared space, allowing their geometric organisation to be examined at 2 levels: global organisation across dependency relations and local lexical neighbourhoods within individual relations. 3 research questions that we will adress are:
\begin{itemize}
    \item Global geometry: What geometric structures emerge when dependency-conditioned lexical units are jointly represented within a shared semantic space?
    \item Local geometry: How are local semantic neighbourhoods organised within the shared semantic space, and to what extent do they reflect relation-specific semantic patterns?
    \item Cross-method comparison: To what extent do count-based and predictive representations exhibit similar global and local geometric structures?
\end{itemize}

\section{Data and Representations}
\subsection{Data}
We used WikiText-103 \cite{merity_pointer_2016}, a large-scale English language modelling corpus comprising approximately 103 million tokens extracted from Wikipedia articles. Its size, linguistic diversity, and public availability make it suitable for training dependency embedding, which requires sufficient observations of lexical items across different syntactic contexts. We pre-processed the data (Appendix~\ref{sec:appendix_data}) then parsed the sentences using Stanza \cite{qi-etal-2020-stanza}.

\subsection{Types of representations}
We construct 4 dependency distributional representations (table \ref{tab:embedding_models})  using 2 count-based and 2 SGNS configurations (original and reversed). In original SGNS, ordinary lemmas serve as targets and dependency-conditioned items as contexts. We additionally consider the reversed orientation, in which dependency-conditioned items become targets and ordinary lemmas become contexts. Since the present study focuses on the geometric organisation of dependency-conditioned lexical units, we examine the representations of these units across all 4 configurations. For count-based models and reversed SGNS, they correspond to target representations, whereas for original SGNS, they correspond to context representations.

Each dependency co-occurrence is represented as a tuple \((A, B/r, f)\), where \(A\) is a lemma, \(B/r\) is a dependency-conditioned lexical unit consisting of a lemma and a dependency relation, and \(f\) is their co-occurrence frequency. To distinguish the direction of dependency relations, we use the prefixes \(\texttt{chi\_}\) and \(\texttt{pa\_}\). For example, if \(A\) is the syntactic head of \(B\) under relation \(r\), the corresponding co-occurrences are \((A, B/\texttt{chi\_}r, f)\) and \((B, A/\texttt{pa\_}r, f)\). In the reversed training configuration, the order of these pairs is inverted, yielding \((B/\texttt{chi\_}r, A, f)\) and \((A/\texttt{pa\_}r, B, f)\), respectively.

\subsubsection{Count-based representations}
We construct 2 count-based models using Positive Pointwise Mutual Information (PPMI), a commonly used association measure in distributional semantics \cite{levy_improving_2015}. Both models use the reversed orientation of the dependency co-occurrence data, such that matrix rows correspond to dependency-conditioned lexical units and columns correspond to ordinary lexical contexts. The constructing procedures are detailed in Appendix~\ref{sec:appendix_count-training}.

\paragraph{PPMI-5000.} The first configuration constructs an explicit sparse representation using the 5000 most frequent context features, ranked according to their marginal frequencies, with ties resolved alphabetically to ensure reproducibility. 

\paragraph{PPMI-SVD.} The second constructs a PPMI matrix using the full context vocabulary and applies Truncated Singular Value Decomposition (SVD) to obtain 300-dimensional vectors.

\subsubsection{SGNS representations}
We employ dependency SGNS \cite{levy_dependency-based_2014}, which extends the original SGNS to support arbitrary context features. We train 2 configurations that differ in the orientation of the target--context pairs. The training procedures are detailed in Appendix~\ref{sec:appendix_count-training}.

\paragraph{Original-context SGNS.} The first configuration follows the original orientation of dependency SGNS, in which ordinary lemmas serve as targets and dependency-conditioned lexical units serve as contexts. The model then learns separate target and context embedding matrices. Although previous dependency embedding studies have primarily examined the target matrix, our analysis focuses on the context matrix, where each dependency-conditioned lexical unit is represented by a 300-dimensional vector.

\paragraph{Reversed SGNS.} The second configuration reverses the orientation of the training pairs, treating dependency-conditioned lexical units as targets and ordinary lemmas as contexts. Consequently, the dependency-conditioned units are learned in the target embedding matrix rather than the context matrix. The resulting 300-dimensional target vectors are extracted for geometric analysis. As reversing the target--context orientation changes the roles of the 2 vocabularies within the SGNS training procedure and negative samples are drawn from a smoothed unigram distribution with an exponent of \(0.75\), reversing the training orientation changes the negative sampling distribution, meaning that the 2 configurations do not necessarily yield equivalent embedding geometries.

\begin{table}[t]
\centering
\small
\setlength{\tabcolsep}{3pt}
\begin{tabularx}{\columnwidth}{
    @{}l>{\raggedright\arraybackslash}Xcc@{}
}
\toprule
Model & Construction & Dim. & Vectors \\
\midrule
PPMI-5000
    & Top 5,000 contexts + PPMI
    & 5,000 & Rows \\
PPMI-SVD
    & Full contexts + PPMI + SVD
    & 300 & Rows \\
Orig. SGNS
    & Lemma $\rightarrow$ lemma/relation
    & 300 & Context \\
Rev. SGNS
    & Lemma/relation $\rightarrow$ lemma
    & 300 & Target \\
\bottomrule
\end{tabularx}
\caption{Summary of the 4 dependency-conditioned
representations. Orig. denotes original-context
SGNS; Rev. denotes reversed SGNS.}
\label{tab:embedding_models}
\end{table}

\section{Experiments and Results}
We investigate the geometric and linguistic organisation of dependency-conditioned  representations at 2 levels: global and local. Global geometry concerns the overall organisation of dependency relations within the shared semantic space, whereas local geometry concerns the relationships among lexical items within individual dependency relations. Unless otherwise specified, cosine similarity is used to measure relationships between embedding vectors. To examine the robustness of the findings, we additionally evaluate the effects of vocabulary size, frequency distribution, and representation method.

\subsection{Global geometry}
We first investigate how dependency relations are organised within the shared semantic space through 2 stages of analysis. To avoid confusion, \texttt{target POS} refers to the POS of the lemma in a dependency-conditioned item, whereas \texttt{context POS} refers to the POS of the lemma syntactically connected to that target.

\begin{figure}[htbp]
    \centering
    \includegraphics[width=0.8\linewidth]{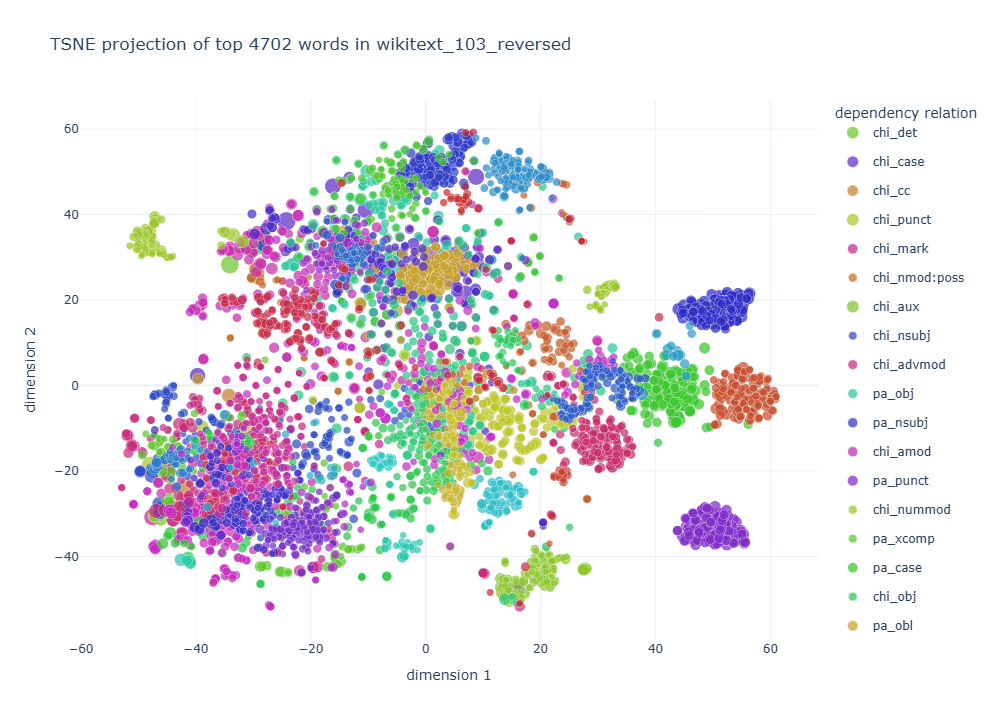}
    \caption{t-SNE projection.}
    \label{fig:TSNE_projection}
\end{figure}

\subsubsection{Exploratory Analysis}

We first explored the global organisation of dependency-conditioned embeddings by selecting the 10000 most frequent lemma--relation types and balancing their representation across dependency relations. We visualised the selected embeddings using t-SNE (Figure~\ref{fig:TSNE_projection}) and applied hierarchical agglomerative clustering (HAC) to relation centroids based on cosine distance (Figure~\ref{fig:relation_centroid_analysis}). The resulting structures suggested systematic groupings of relations associated with similar contextual POS distributions.

\begin{figure}[htbp]
    \centering
    \includegraphics[width=0.8\linewidth]{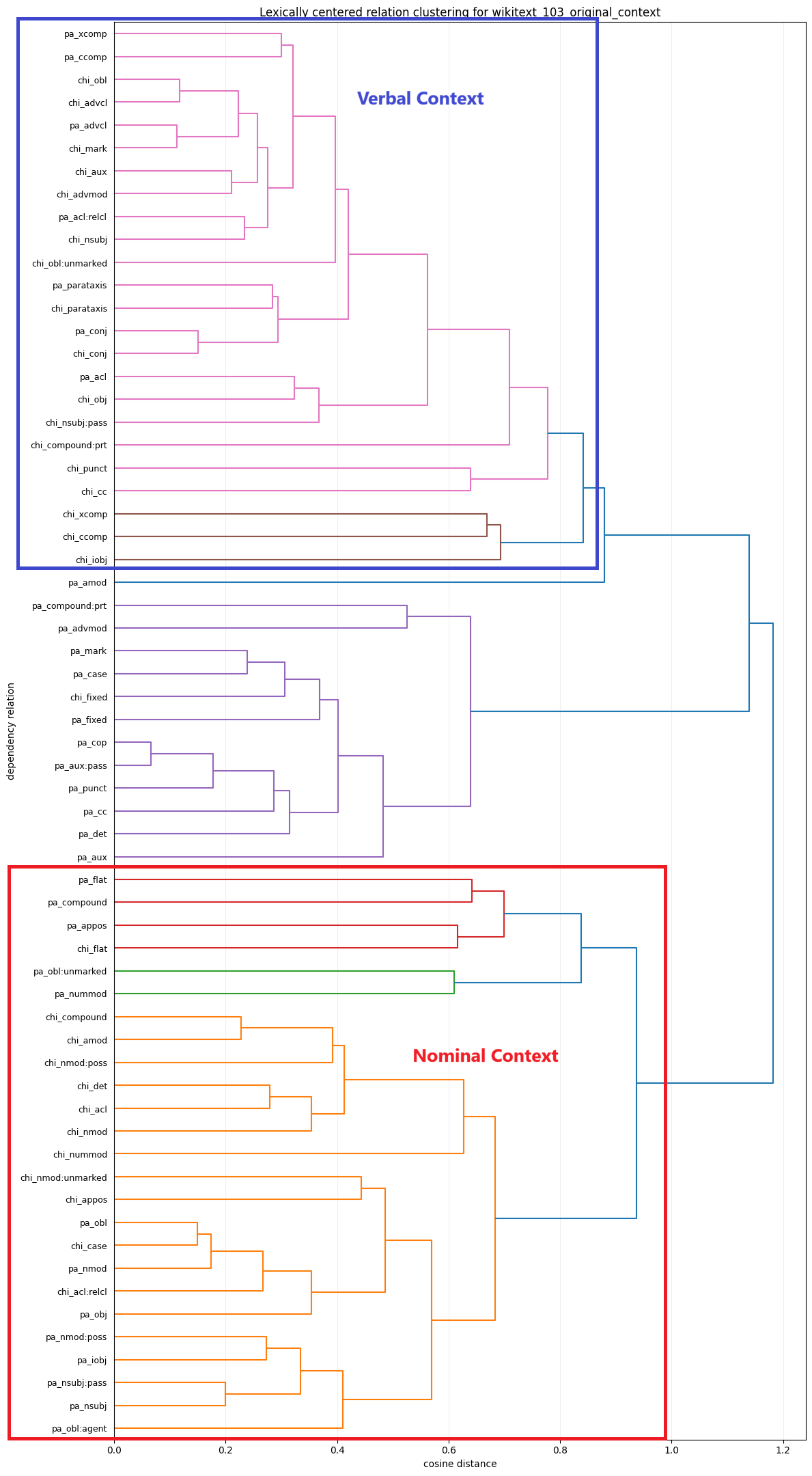}
    \caption{HAC dendrogram for lexically centred relation analysis}
    \label{fig:lexically_centred_relation_analysis}
\end{figure}

To examine whether these patterns were driven by differences in lexical composition, we conducted a second exploratory analysis using within-lemma centring. We subtracted each lemma's mean vector across its available relation variants before computing residual relation centroids and applying HAC (Figure~\ref{fig:lexically_centred_relation_analysis}). Similar groupings still existed and persisted across all 4 representations, suggesting that the observed organisation was not solely attributable to lemma-specific embedding positions. Detailed sampling procedures, centring equations, and clustering settings are provided in Appendix~\ref{sec:appendix_explore_global}.

These observations motivated a quantitative analysis of the association between relation-level embedding geometry and POS distributions.

\subsubsection{POS distribution and embedding geometry}
To quantify the patterns observed in the exploratory analyses, we examined whether \textit{dependency relations that are geometrically similar also have similar POS distributions}. Specifically, we compared the \textbf{distances between lexically centred relation centroids} with the \textbf{differences in the POS distributions} of their represented lemmas (target POS) and syntactic counterparts (context POS).

For each relation, POS distributions were reconstructed from the parsed corpus, normalised for each \texttt{lemma-relation} type, and then averaged across types. We measured embedding distances using cosine distance and POS-distribution differences using Jensen--Shannon distance. Spearman's rank correlation was used to assess the correspondence between these distances across relation pairs. Statistical significance was evaluated using 10,000 relation-label permutations. We additionally tested whether context-POS alignment
was stronger than target-POS alignment ($\Delta\rho = \rho_{\mathrm{context}} - \rho_{\mathrm{target}}$).

\begin{table}[htbp]
\centering

\caption{Correlations between lexically centred
relation distances and POS-distribution distances
at the top-10000 vocabulary cutoff.}
\label{tab:correlations_lexically_centred_POS_dist}
\small
\setlength{\tabcolsep}{4pt}
\begin{tabular}{lccc}
\hline
\textbf{Model} &
\textbf{Ctx. $\rho$} &
\textbf{Tgt. $\rho$} &
\textbf{$\Delta\rho$} \\
\hline
PPMI-SVD
  & $.603^{***}$ & $-.128^{***}$ & $.731^{***}$ \\
Orig. SGNS
  & $.576^{***}$ & $-.100^{**}$ & $.676^{***}$ \\
PPMI-5000
  & $.575^{***}$ & $-.112^{***}$ & $.687^{***}$ \\
Rev. SGNS
  & $.463^{***}$ & $-.094^{**}$ & $.557^{***}$ \\
\hline
\end{tabular}

\vspace{2pt}
\parbox{\columnwidth}{%
\footnotesize
\textit{Note:} Ctx. = Context-POS;
Tgt. = Target-POS;
Orig. = Original-context;
Rev. = Reversed.
$\Delta\rho = \rho_{\mathrm{ctx}} -
\rho_{\mathrm{tgt}}$.
$^{**}p < .01$;
$^{***}p < .001$
(10000 relation-label permutations).
Correlations were tested two-sided;
$\Delta\rho$ was tested one-sided.
}
\end{table}

\paragraph{Context-POS alignment}
As shown in Table~\ref{tab:correlations_lexically_centred_POS_dist}, all 4 representations exhibited significant positive correlations between relation-centroid distances and context-POS distances (\(\rho = .463\text{--}.603\), \(p < .001\)). In contrast, correlations with target-POS distances were weakly negative (\(\rho = -.094 \text{ to } -.128\)). Direct permutation tests confirmed that context-POS alignment was significantly stronger than target-POS alignment in all 4 representations (\(\Delta\rho = .557\text{--}.731\), \(p < .001\)). These results support the hypothesis that the global geometry of lexically centred dependency-relation embeddings is more closely associated with the POS composition of context words than with that of the embedded lexical items.

\paragraph{Robustness and frequency sensitivity.}
To assess whether context-POS alignment depends on vocabulary selection, we repeated the analysis across 9 cumulative vocabulary cutoffs (top 200--20,000). At every cut-off, across all 4 representations, context-POS correlations remained positive and significant ($p<.001$), and direct
context-minus-target comparisons were also significant ($p<.001$) (Table~\ref{tab:context_pos_sensitivity}). This indicates that
the main finding is not specific to the primary top-10,000 selection. However, because cumulative vocabulary expansion also increases the number of retained dependency relations, variation in correlation magnitude may reflect changes in both lexical coverage and relation composition. Detailed results are provided in
Appendix~\ref{sec:appendix_POS_dist_global}.

To investigate whether context-POS alignment varied across lexical frequency ranges, we conducted a second sensitivity analysis using 6 non-overlapping frequency bands using a fixed set of 29 dependency relations shared by all bands. Context-POS alignment remained positive and significant across all bands and representations ($p<.001$;
Table~\ref{tab:context_pos_bands}). However, different models exhibited different descriptive frequency profiles. In particular, the count-based representations exhibited stronger alignment among higher-frequency units, whereas reversed SGNS showed stronger alignment among lower-frequency units.

These quantitative findings support the context-POS organisation suggested by the HAC analyses. It is robust across vocabulary selections, although its strength varies with lexical frequency and representation method.

\subsection{Local Geometry}
\label{sec:local_geometry}
We investigate local lexical geometry from 3 complementary perspectives: its consistency across dependency relations, its agreement across embedding methods, and its linguistic interpretability. To ensure comparability across vocabulary sizes, we retained 112 dependency-relation pairs with at least 2,000 shared lemmas. The same relation pairs were used across all vocabulary cutoffs to prevent changes in relation-pair coverage from affecting the comparisons. Further details on shared-vocabulary coverage and pair selection are provided in Appendix~\ref{sec:appendix_shared_vocab}.

\subsubsection{Cross-Relation Geometric Consistency}
\label{sec:cross_relation_geometry}

\paragraph{Shared-vocabulary geometric consistency.}
To assess whether lexical similarity structures are preserved across dependency relations, we selected \(N\) shared lemmas for each of the 112 eligible relation pairs. Within each relation, we calculated pairwise cosine similarities among the selected lemmas and compared their relative ordering across relations using Spearman's rank correlation. The resulting correlations were macro-averaged across relation pairs. Higher values indicate greater consistency in lexical similarity rankings across dependency relations. The sampling procedures are provided in Appendix~\ref{sec:appendix_cross_relation}.

Interestingly, PPMI-based representations consistently exhibited higher cross-relation correlations than SGNS representations (Table~\ref{tab:cross_space_correlations}). This pattern was observed across vocabulary sizes, indicating greater preservation of relative lexical similarity rankings across dependency relations in the PPMI-based representations.

\paragraph{Local neighbourhood consistency.}
While pairwise similarity correlations assess the preservation of lexical similarity rankings, they
do not indicate whether individual lemmas retain the same nearest neighbours. We therefore compare
the top-10 nearest-neighbour sets of each shared lemma across dependency relations using chance-corrected Jaccard overlap: Scores are averaged across lemmas within each relation pair and then macro-averaged across the same 112 relation pairs. Further implementation details and
vocabulary sensitivity results are provided in Appendix~\ref{sec:appendix_cross_relation}.

Neighbourhood overlap remains low across all 4 representations (Table~\ref{tab:knn_jaccard}). Nevertheless, PPMI models consistently yield the higher Jaccard scores than SGNS models, further supporting the observation that PPMI-based representations exhibit greater cross-relation consistency in both lexical similarity rankings and immediate neighbourhoods.

\begin{table}[htbp]
\centering
\caption{Macro-mean chance-corrected KNN Jaccard
across 112 dependency relation pairs.}
\label{tab:knn_jaccard}
\small
\setlength{\tabcolsep}{5pt}
\begin{tabular}{rcccc}
\hline
\textbf{$N$} &
\textbf{Rev.} &
\textbf{Orig.} &
\textbf{SVD} &
\textbf{PPMI} \\
\hline
200   & .080 & .088 & .127 & .143 \\
500   & .073 & .079 & .107 & .124 \\
1000  & .065 & .070 & .095 & .112 \\
1100  & .063 & .067 & .092 & .108 \\
1200  & .061 & .065 & .090 & .106 \\
1500  & .057 & .061 & .084 & .099 \\
2000  & .053 & .056 & .078 & .091 \\
\hline
\end{tabular}

\vspace{2pt}
\parbox{\columnwidth}{%
\footnotesize
\textit{Note:} Rev. = Reversed SGNS;
Orig. = Original-context SGNS;
SVD = PPMI-SVD (300d);
PPMI = PPMI-5000.
}
\end{table}

Together, these findings suggest that dependency-conditioned embeddings preserve some broader lexical similarity regularities across dependency relations, while exhibiting substantial variation in their immediate lexical neighbourhoods. The extent of this cross-relation consistency also depends on the representation method, with PPMI-based representations showing greater consistency than SGNS-based representations.

\subsubsection{Cross-Method Comparison}
\label{sec:cross_method_comparison}

We further examine the lexical structures within individual dependency relations from 2 complementary perspectives: their correspondence with human semantic judgements and their consistency across representation methods.

\paragraph{Lexical Semantic Evaluation.}
We first evaluated all representations against the WordSim-353 Similarity and Relatedness subsets. Across 14 dependency relations with matched benchmark coverage, correlations with human judgements were consistently higher for Similarity than for Relatedness. PPMI-5000 achieved the highest correlations for 9 of 14 relations on Similarity and 7 of 14 on Relatedness. These results suggest that the representations capture lexical similarity information reflected in human judgements, with stronger alignment to similarity than to relatedness ratings. Moreover, the performance of PPMI-5000 indicates that predictive representations are not necessarily superior to count-based alternatives on this benchmark, consistent with previous comparative studies \cite{lenci_comparative_2022, levy_improving_2015}.  A particularly notable result concerns the 2 SGNS representations. Original-context SGNS outperforms Reversed SGNS for 9/14 relations on Similarity and 10/14 relations on Relatedness, even though reversed SGNS is trained specifically to represent dependency-conditioned units as target vectors. Thus, the original SGNS context vectors, which are learned jointly with the target vectors but are often treated as a secondary training output and discarded, retain more benchmark-relevant semantic information for most dependency relations. Detailed evaluation procedures and results are provided in Appendix~\ref{sec:appendix_wordsim}.

\paragraph{Cross-Method Representational Similarity Analysis (RSA).}
To examine whether different embedding methods preserve similar lexical structures within individual dependency relations, we compared pairwise cosine similarity rankings across the 4 representations. For each of 50 dependency relations, we selected 200 shared lexical units and calculated Spearman correlations between their similarity matrices. Agreement was higher within methodological families (mean $\rho=.774$--$.791$) than between PPMI and SGNS (mean $\rho=.581$--$.654$). Notably, cross-family agreement varied substantially across dependency relations, from $\rho=.911$ for \texttt{pa\_conj} to $\rho=-.029$ for \texttt{pa\_det}. These findings suggest that the extent to which lexical geometry is preserved across methods depends on the dependency relation. Full RSA results are provided in Appendix~\ref{sec:appendix_rsa}.

Overall, the findings of this section suggest that dependency-conditioned representations capture meaningful lexical semantic structures, while their geometric organisation varies across both dependency relations and representation methods.

\subsubsection{Linguistic Interpretation}
\label{sec:linguistic_interpretation}

\paragraph{Within-relation lexical structure}
We qualitatively inspect the dendrograms of several relations (but use only \texttt{pa\_nsubj} of Original-context SGNS for illustration) to examine whether lemma--relation embeddings belonging to the same dependency relation exhibit meaningful internal organisation. Of the 2,000 most frequent lemma--relation types, 108 belong to \texttt{pa\_nsubj}. Figure~\ref{fig:pa_nsubj_dendrogram} reveals several semantically coherent branches, including cognitive and attitudinal verbs (\textit{think--feel, consider--believe, decide--want}), communication verbs (\textit{describe--say, claim--suggest--argue--tell}), and process and event verbs (\textit{begin--start, result--cause--occur, release--announce}), alongside verbs of composition (\textit{consist--contain--include}).. This therefore suggests that a single dependency-conditioned space can preserve fine-grained lexical structure rather than forming an undifferentiated cloud. Across the 4 models, several local lexical-semantic groupings remain relatively consistent, whereas their integration into larger hierarchical clusters varies across embedding methods.

\paragraph{Nearest-neighbour exploration}
To explore relation-specific semantic differentiation, we manually selected 30 English lemmas exhibiting within-POS polysemy or cross-POS ambiguity, each represented under multiple dependency relations. We compared their nearest-neighbour structures across relations to examine how different syntactic contexts highlight distinct meanings or usages. The full candidate list is provided in Table~\ref{tab:polysemy_candidates}.

For each query \texttt{lemma/relation} (e.g., \texttt{bank/chi\_compound}), we compute cosine similarity against all non-zero candidate vectors carrying the same relation suffix and retain the 5 candidates with the highest cosine similarity. Restricting the candidate set to the same relation ensures that a neighbourhood compares lexical items within 1 relation-conditioned space. Table~\ref{tab:within_pos_examples} reports 6 Within-POS cases, in which \textit{bank}, \textit{hold}, and \textit{hard} retain their broad \texttt{Noun}, \texttt{Verb}, and \texttt{Adjective} status, respectively, while different relations emphasise different readings within that category. \textit{bank} provides the clearest nominal example. Under \texttt{chi\_compound}, its neighbourhood is predominantly financial or institutional across all 4 spaces, with neighbours such as \textit{insurance, mortgage, bankruptcy}. Under \texttt{chi\_obl}, all 4 spaces shift towards the geographical river-bank reading, retrieving \textit{tributary, estuary, valley}. The precise neighbours vary, but the relation-conditioned sense contrast is stable across training algorithms. \textit{hold} illustrates a frame-sensitive verbal contrast. Its \texttt{chi\_acl} representation is surrounded by verbs associated with hosting or conducting events, including \textit{host, contest, preside}. Its \texttt{pa\_ccomp} representation instead lies near verbs of propositional stance and communication, including \textit{argue, maintain, conclude, believe}. The dependency relation therefore changes which facet of the same lemma dominates its local geometry. \textit{hard} shows particularly strong cross-model agreement. The \texttt{chi\_amod} neighbourhood emphasises physical or material properties through \textit{soft, tough, solid, coarse}, and related descriptors. In contrast, \texttt{pa\_ccomp} consistently retrieves \textit{difficult, easy, impossible}, highlighting an evaluative judgement about task difficulty or possibility. This contrast appears in both SGNS and PPMI spaces despite differences in the remaining neighbours. The selected Within-POS examples therefore exhibit broadly consistent semantic contrasts across the 4 representations, despite differences in individual nearest neighbours.

\begin{table*}[t]
\centering
\caption{Within-POS examples. 5 neighbours are shown in decreasing cosine-similarity order.}
\label{tab:within_pos_examples}
\scriptsize
\setlength{\tabcolsep}{3pt}
\begin{tabular}{p{1.8cm} l p{2.65cm} p{2.65cm} p{2.65cm} p{2.65cm}}
\hline
\textbf{Reading} &
\textbf{Relation} &
\textbf{Reversed SGNS} &
\textbf{Original-context SGNS} &
\textbf{PPMI-SVD} &
\textbf{PPMI-5000} \\
\hline
bank: financial &
\texttt{chi\_compound} &
insurance, mortgage, booking, government, bankruptcy &
mortgage, gate, booking, wire, KGB &
government, service, property, estate, council &
government, banking, service, company, Bank \\
\hline
bank: river edge &
\texttt{chi\_obl} &
tributary, estuary, foothill, valley, ravine &
valley, estuary, Plain, Firth, shore &
hill, river, shore, valley, River &
River, town, area, shore, south \\
\hline
hold: event hosting &
\texttt{chi\_acl} &
take, contest, host, preside, benefit &
organised, contest, host, preside, take &
host, organised, attend, organize, celebrate &
take, host, attend, organised, call \\
\hline
hold: propositional stance &
\texttt{pa\_ccomp} &
argue, establish, conclude, maintain, believe &
argue, rule, maintain, conclude, assert &
rule, contend, maintain, determine, establish &
argue, believe, suggest, claim, state \\
\hline
hard: physical property &
\texttt{chi\_amod} &
soft, process, tougher, shiny, tough &
tough, tougher, slippery, tinged, flavor &
soft, solid, melt, loose, coarse &
soft, tough, solid, more, color \\
\hline
hard: difficulty judgement &
\texttt{pa\_ccomp} &
difficult, easy, impossible, affect, hope &
easy, impossible, difficult, discuss, affect &
difficult, impossible, easy, help, teach &
difficult, easy, impossible, state, say \\
\hline
\end{tabular}
\end{table*}

\begin{table*}[t]
\centering
\caption{Cross-POS examples. 5 neighbours are shown in decreasing cosine-similarity order.}
\label{tab:cross_pos_examples}
\scriptsize
\setlength{\tabcolsep}{3pt}
\begin{tabular}{p{1.8cm} l p{2.65cm} p{2.65cm} p{2.65cm} p{2.65cm}}
\hline
\textbf{Reading} &
\textbf{Relation} &
\textbf{Reversed SGNS} &
\textbf{Original-context SGNS} &
\textbf{PPMI-SVD} &
\textbf{PPMI-5000} \\
\hline
file: document &
\texttt{chi\_obj} &
e-mail, diary, inventory, email, Spirit &
diary, gem, software, poison, Spirit &
sample, fragment, photograph, tape, photo &
document, data, information, map, tape \\
\hline
file: submit or lodge &
\texttt{pa\_obj} &
lodge, print, forward, invalidate, submit &
lodge, forward, retract, circulate, rescind &
issue, forward, submit, retract, appeal &
lodge, submit, withdraw, dismiss, issue \\
\hline
face: body part &
\texttt{chi\_obj} &
breast, sand, fence, cylinder, trunk &
breast, throat, paint, rock, sand &
eye, hair, head, breast, tail &
eye, head, nose, hand, image \\
\hline
face: confront &
\texttt{pa\_obj} &
overcome, fear, eliminate, avoid, over &
over, defeat, eliminate, confront, overcome &
eliminate, defeat, beat, pit, challenge &
defeat, beat, avoid, eliminate, overcome \\
\hline
fine: quality &
\texttt{chi\_amod} &
quality, coarse, superb, exquisite, elegant &
exquisite, elegant, quality, pleasing, colourful &
polished, brilliant, elegant, soft, stain &
good, famous, excellent, brilliant, beautiful \\
\hline
fine: financial penalty &
\texttt{chi\_obj} &
dividend, subsidy, Assent, penalty, tax &
dividend, Assent, condemnation, bribe, pension &
tax, fee, sanction, suspension, writ &
penalty, tax, ban, fee, sanction \\
\hline
present: current &
\texttt{chi\_amod} &
current, new, different, same, pre-exist &
current, serve, convenient, pre-exist, new &
current, certain, typical, same, particular &
current, new, different, certain, exist \\
\hline
present: model-dependent verbal neighbourhood &
\texttt{pa\_obj} &
see, compare, find, tout, view &
see, compare, address, find, accept &
view, consider, compare, base, characterize &
consider, describe, introduce, show, discuss \\
\hline
\end{tabular}
\end{table*}

Table~\ref{tab:cross_pos_examples} reports 8 Cross-POS cases. \textit{file} and \textit{face} contrast nominal and verbal readings, while \textit{fine} and \textit{present} contrast adjective-centred readings with nominal or verbal alternatives. Cross-POS differentiation is also consistent in several cases. \texttt{file/chi\_obj} is organised around documents, data, photographs, and recorded information, whereas \texttt{file/pa\_obj} is surrounded by submission verbs such as \textit{submit, retract, withdraw}; \texttt{face/chi\_obj} retrieves body-part terms such as \textit{eye, head, nose}, while \texttt{face/pa\_obj} retrieves confrontational actions such as \textit{confront, overcome, defeat}; \texttt{fine/chi\_amod} expresses positive quality through \textit{superb, exquisite, elegant}; \texttt{fine/chi\_obj} instead emphasises penalties through \textit{tax, fee, sanction}. These contrasts are stable across all 4 models and are therefore interpreted once as shared cross-model patterns. The Cross-POS cases also reveal a clearer division in model strengths. The PPMI spaces often provide the most transparent category separation: PPMI-SVD and PPMI-5000 produce especially coherent nominal and verbal sets for file, face, fine. The SGNS spaces are strong when the alternative reading is tied to a selective verbal frame, as in \texttt{file/pa\_obj}, \texttt{face/pa\_obj}, but their nominal neighbourhoods more often contain specialised or idiosyncratic items. \texttt{present/pa\_obj} is the clearest cross-model disagreement and is therefore analysed separately. The difference is interpretive: the first 3 spaces foreground evaluation or perception, whereas PPMI-5000 foregrounds the act of presenting information.

In the selected Within-POS examples, the 4 models generally exhibit similar relation-conditioned semantic interpretations, with differences primarily concerning lexical specificity. The Cross-POS examples also show interpretable differentiation, but reveal greater variation in neighbourhood coherence and semantic emphasis across representations.

\section{Discussion and Conclusion}

Our study provides a systematic characterisation of the linguistic and geometric organisation of dependency-conditioned lexical representations within a shared semantic space. Three main findings emerge. First, at the global level, dependency-relation geometry is systematically associated with the POS composition of lexical contexts rather than that of the represented lemmas. This pattern persists after within-lemma centring and is consistently observed across all representations, vocabulary selections, and frequency bands. Second, at the local level, dependency relations exhibit distinct but partially overlapping lexical similarity structures. Pairwise similarity correlations reveal some preservation of broader lexical relationships across relations, whereas relatively low nearest-neighbour overlap indicates substantial variation in immediate lexical neighbourhoods. Qualitative analyses further show that these neighbourhoods can highlight different meanings or usages of polysemous words, consistent with earlier observations \cite{mouton_unsupervised_2009,grefenstette_conquering_2007}. Third, geometric organisation is sensitive to the embedding method, particularly at the local level. PPMI-based representations exhibit greater cross-relation consistency in both pairwise similarity rankings and nearest-neighbour overlap than SGNS representations. Cross-method RSA further reveals stronger agreement within methodological families than between count-based and predictive models, with substantial variation across dependency relations. This suggests that the extent to which lexical geometry is preserved across methods is itself relation-dependent. Taken together, these findings suggest that dependency-conditioned semantic spaces exhibit complementary forms of linguistic organisation at different geometric levels: systematic global regularities associated with contextual POS distributions alongside locally differentiated lexical structures. Although these patterns emerge across all four representations, their geometric consistency varies across embedding methods. Rather than treating lemma--relation vectors solely as contextual features, our results highlight the value of examining their shared semantic space as a structured representation in its own right. This multi-level perspective provides a basis for interpreting and selecting dependency-conditioned representations for lexical semantic research.

\section{Limitation}
This study has 2 main limitations. First, the semantic evaluation of cross-relation similarities remains limited. Although \citet{baroni_distributional_2010} provided an initial demonstration of their linguistic utility through verb classification, systematic semantic evaluation remains unexplored. Second, the polysemy analysis relies on manually selected examples rather than systematic quantitative evaluation. Both limitations stem from the lack of dedicated benchmarks for assessing semantic similarities and distinctions between dependency-conditioned lexical units, both within and across POS categories.


\bibliography{custom}

\appendix

\section{Appendix A - Data Processing and Embedding Implementation Details}

\subsection{Data Processing}
\label{sec:appendix_data}
The WikiText-103 corpus was divided into articles, stripped of headings and section markup, normalised for Unicode and HTML entities, and corrected for WikiText-specific tokenisation artefacts. The text was
detokenised while preserving case and punctuation, segmented into complete sentences, and processed with Stanza for tokenisation, lemmatisation, POS tagging, and dependency parsing.

\subsection{Count-based Representation}
\label{sec:appendix_count-training}
Given a co-occurrence matrix \(M\), the association between a target \(w\) and a context \(c\) is calculated as:

\begin{equation}
\operatorname{PPMI}(w,c) = \max\left(0, \log\frac{P(w,c)}{P(w)P(c)}\right)
\end{equation}

where the probabilities are estimated from the co-occurrence counts in the corresponding matrix. Negative PMI values are replaced with zero, producing non-negative association weights. Note that we only retain dependency-conditioned items whose marginal frequency is at least 100 to ensure stability of the semantic space.

Both count-based configurations used dependency-conditioned units as rows. PPMI-5000 retained the 5000 context lemmas with the highest marginal frequencies (alphabetical tie-breaking); the PPMI probabilities were estimated after restricting the count matrix to those
columns. PPMI-SVD instead computed PPMI over the full available context vocabulary before applying truncated SVD to 300 dimensions. Dependency-conditioned units with marginal frequency below 100 were excluded.

\subsection{SGNS representations}
\label{sec:appendix_sgns-training}
Both SGNS configurations used \texttt{word2vecf} with 300 dimensions, 15 negative samples, 5 iterations, and 10 threads. We first restricted the reversed co-occurrence counts to dependency-conditioned units with marginal frequency at least 100, recalculated both vocabulary marginals on the retained pairs, and expanded counts into repeated training events. The same retained pairs were used in both configurations: original-context SGNS treated lemmas as targets and lemma–relation units as contexts, whereas reversed SGNS swapped these roles. Negative samples followed a smoothed unigram distribution with exponent 0.75.

\section{Appendix B - Global Geometry Experiment}
\subsection{Exploratory Analysis for Global Structure}
\label{sec:appendix_explore_global}

\subsubsection{Relation centroid analysis}
\label{sec:relation_centroid_analysis}
We ranked lemma–relation types by frequency and selected the top 10,000. These types spanned 73 of the 84 directed relation labels, giving a cap of $\lfloor 10000 / 73 \rfloor = 136$ types per relation. Relations with fewer than 10 units are excluded, while each remaining relation contributes up to its 136 highest-frequency types. This
yielded 4,751 types, of which 4,702 remained after restricting lemmas to those represented in at least 2 relations. We used t-SNE for an illustrative projection of the selected vectors.

\begin{figure}[htbp]
    \centering
    \includegraphics[width=0.8\linewidth]{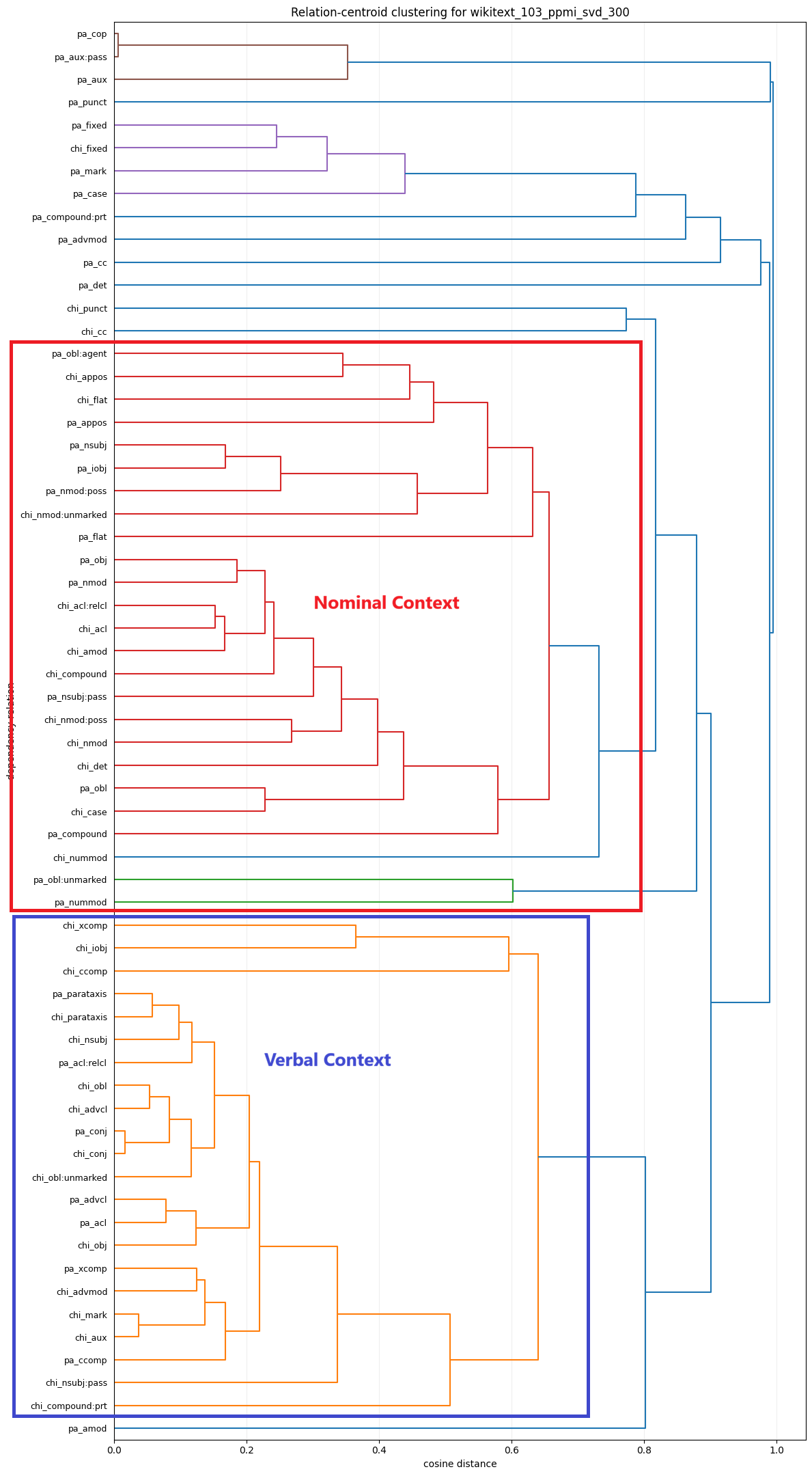}
    \caption{HAC dendrogram for relation centroid analysis}
    \label{fig:relation_centroid_analysis}
\end{figure}

We then calculate the mean embedding vector for each dependency relation and examine the cosine similarities between the resulting relation centroids through HAC. In Figure~\ref{fig:relation_centroid_analysis}, there is a clear region where the POS of the context is predominantly Noun-like and another region where the POS of the context is predominantly Verb-like. This pattern is consistent across different semantic spaces.

\subsubsection{Lexically centred relation analysis}
Starting from the lemma–relation types selected in section~\ref{sec:relation_centroid_analysis}, we retrieved all available relation variants for each lemma from the full embedding, retaining only lemmas represented in at least 2 relations as lemmas occurring in only 1 relation would yield 0 residual vectors after centring. 

For a lemma \(w\), its baseline was calculated over all available relations \(D_w\):
\[ \boldsymbol{\mu}_w=\frac{1}{|D_w|}\sum_{d\in D_w}\mathbf{v}_{w,d} \]
We then subtracted this baseline from each selected lemma–relation vector to obtain a residual indicating its displacement from the lemma's average position:
\[ \mathbf{r}_{w,d}=\mathbf{v}_{w,d}-\boldsymbol{\mu}_w \]
Finally, for each dependency relation \(d\), we averaged the residuals of its selected lemmas \(S_d\) to obtain a relation-level displacement vector:
\[ \mathbf{c}_d=\frac{1}{|S_d|}\sum_{w\in S_d}\mathbf{r}_{w,d} \]
Cosine distances between these centroids were used for HAC (Figure~\ref{fig:lexically_centred_relation_analysis}). Similar centroid directions indicate similar average within-lemma shifts across relations.

\subsection{POS Distributions and Embedding Geometry}
\label{sec:appendix_POS_dist_global}

\begin{table}[htbp]
\centering
\caption{Context-POS alignment across 9 cumulative vocabulary cutoffs.}
\label{tab:context_pos_sensitivity}
\begin{tabular}{lc}
\hline
\textbf{Representation} & \textbf{Context-POS $\rho$} \\
\hline
PPMI-SVD          & $.577\text{--}.796$ \\
PPMI-5000        & $.560\text{--}.651$ \\
Original-context SGNS & $.545\text{--}.749$ \\
Reversed SGNS         & $.405\text{--}.494$ \\
\hline
\end{tabular}

\vspace{0.5em}
\begin{minipage}{\linewidth}
\footnotesize
\textit{Note.} All reported context-POS correlations and direct context-minus-target comparisons were significant at \(p < .001\). The number of retained relations increased from 16 at the top 200 cutoff to 65 at the top 20000 cutoff.
\end{minipage}
\end{table}

\newpage
\begin{table}[htbp]
\centering
\caption{Context-POS correlations across 6
non-overlapping frequency bands, using a fixed
set of 29 dependency relations.}
\label{tab:context_pos_bands}
\small
\setlength{\tabcolsep}{4pt}
\begin{tabular}{lcccc}
\hline
\textbf{Freq. ranks} &
\textbf{Rev.} &
\textbf{Orig.} &
\textbf{SVD} &
\textbf{PPMI} \\
\hline
1--1000        & .420 & .621 & .738 & .605 \\
1001--3000     & .447 & .612 & .714 & .647 \\
3001--10000    & .450 & .562 & .632 & .617 \\
10001--30000   & .479 & .546 & .539 & .563 \\
30001--60000   & .520 & .527 & .438 & .458 \\
60001--120000  & .609 & .539 & .368 & .390 \\
\hline
\end{tabular}

\vspace{2pt}
\parbox{\columnwidth}{%
\footnotesize
\textit{Note:} Rev. = Reversed SGNS;
Orig. = Original-context SGNS;
SVD = PPMI-SVD (300d);
PPMI = PPMI-5000.
Values are Spearman correlations between
lexically centred relation distances and
context-POS distances.
All correlations and direct context-minus-target
comparisons were significant at $p < .001$
(10000 permutations).
}
\end{table}

\section{Appendix C - Local Geometry Experiment}

\subsection{Shared-Vocabulary Coverage and Relation-Pair Selection}
\label{sec:appendix_shared_vocab}
\begin{figure}[htbp]
    \centering
    \includegraphics[width=0.8\linewidth]{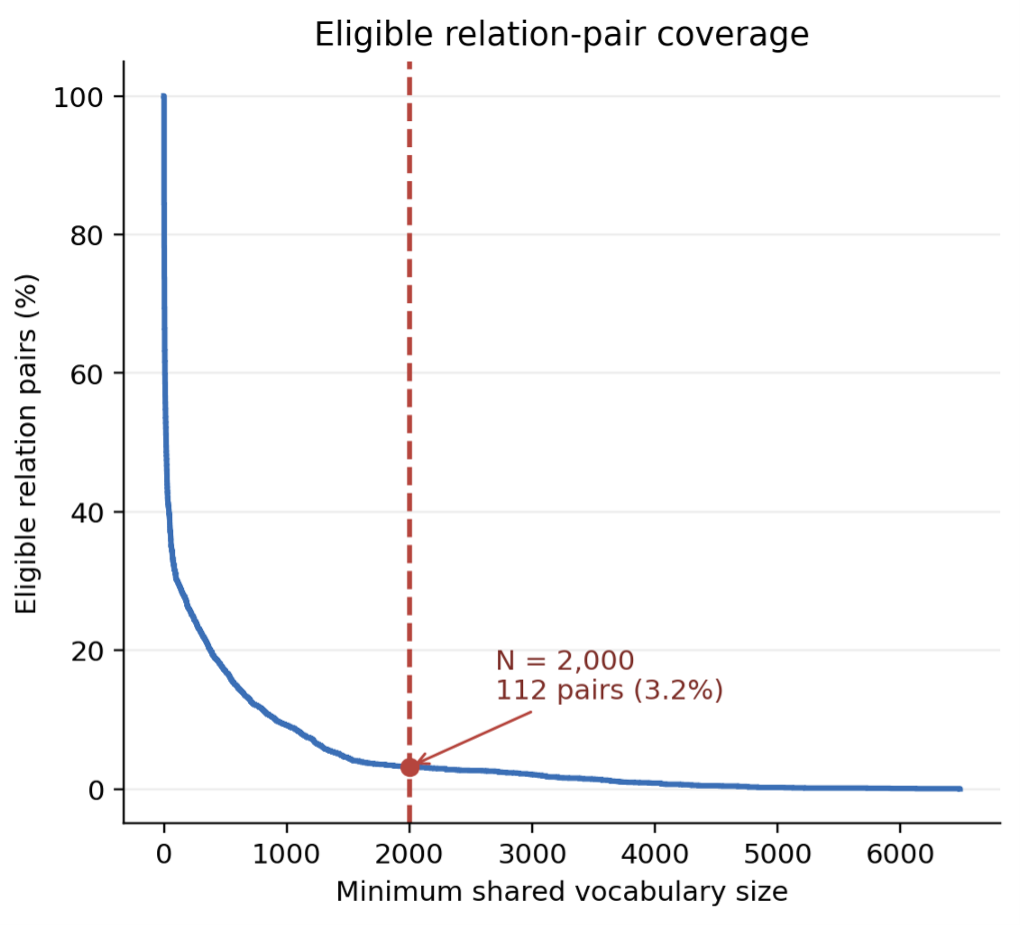}
    \caption{Distribution of shared-lemma counts and the percentage of relation pairs retained at each minimum vocabulary size.}
    \label{fig:shared_vocab}
\end{figure}

The 84 directed dependency-relation labels yield $\binom{84}{2} = 3,486$ possible unordered relation pairs. However, most relation pairs share too few usable lemmas to support comparisons over large vocabularies (figure~\ref{fig:shared_vocab}). We decided to retain relation pairs with at least 2000 shared lemmas, producing a fixed set of 112 relation pairs.

\subsection{Cross-Relation Geometric Consistency}
\label{sec:appendix_cross_relation}

\paragraph{Shared-vocabulary geometric consistency}
We identify lemmas shared by each eligible pair of relations, then rank the shared lemmas by aggregated corpus frequency and retain the requested number \(N\). Within each relation, pairwise cosine similarities are calculated for all selected lemmas. Spearman correlation is then used to compare the resulting similarity structures between relations. Higher correlations indicate greater consistency, whereas lower correlations suggest stronger geometric reorganisation across dependency relations. We repeat the calculation at N = 10, 20, 50, 100, 200, 500, 1000, 1100, 1200, 1500, 2000, while retaining the same 112 relation pairs. The reported cross-space values in Table~\ref{tab:cross_space_correlations} are macro-averages of the 112 relation-pair correlations.

\begin{table}[htbp]
\centering
\caption{Macro-mean Spearman correlations across
112 dependency relation pairs.}
\label{tab:cross_space_correlations}
\small
\setlength{\tabcolsep}{5pt}
\begin{tabular}{rcccc}
\hline
\textbf{$N$} &
\textbf{Rev.} &
\textbf{Orig.} &
\textbf{SVD} &
\textbf{PPMI} \\
\hline
10   & .083 & .109 & .252 & .332 \\
20   & .156 & .191 & .352 & .435 \\
50   & .150 & .195 & .326 & .427 \\
100  & .154 & .185 & .354 & .459 \\
200  & .153 & .183 & .360 & .465 \\
500  & .163 & .191 & .374 & .465 \\
1000 & .165 & .195 & .376 & .450 \\
1100 & .164 & .195 & .373 & .444 \\
1200 & .164 & .195 & .370 & .439 \\
1500 & .165 & .197 & .364 & .425 \\
2000 & .174 & .206 & .357 & .408 \\
\hline
\end{tabular}

\vspace{2pt}
\parbox{\columnwidth}{%
\footnotesize
\textit{Note:} Rev. = Reversed SGNS;
Orig. = Original-context SGNS;
SVD = PPMI-SVD (300d);
PPMI = PPMI-5000.
}
\end{table}

\paragraph{Local neighbourhood consistency}
For each eligible relation pair, we select the same frequency-ranked shared vocabulary of \(N\) lemmas. Each lemma is treated as a query, and its 10 nearest neighbours are retrieved independently under the 2 relations. We then calculate the Jaccard overlap between the resulting neighbour sets. Because some overlap can occur by chance, particularly when the candidate vocabulary is small, we estimate the expected overlap between 2 random sets of 10 neighbours drawn from the same \(N-1\) candidates and apply the correction \((J_{\mathrm{observed}}-J_{\mathrm{expected}})/(1-J_{\mathrm{expected}})\). A corrected value of 0 indicates chance-level overlap, whereas a value of one indicates identical neighbour sets. The corrected scores are first averaged across query lemmas within each relation pair and subsequently macro-averaged across the fixed set of 112 relation pairs.

We repeat this procedure at N = 200, 500, 1000, 1100, 1200, 1500, 2000. Smaller vocabulary sizes are excluded because a neighbour set of 10 represents a substantial proportion of the available candidates (for example, approximately 10\% at \(N=100\)). This restricted candidate pool mechanically increases the probability of neighbour overlap and makes the Jaccard measure less discriminating, even after chance correction. Beginning at \(N=200\), the 10 neighbours constitute only about 5\% of the candidate vocabulary, providing a more informative assessment of local-neighbourhood preservation. The macro-mean chance-corrected KNN Jaccard across 112 relation pairs are reported in table~\ref{tab:knn_jaccard}. 

\subsection{Cross-Method Comparison}
\label{sec:appendix_cross_method}

\paragraph{Lexical semantic evaluation}
\label{sec:appendix_wordsim}
The WordSim-353 Evaluation is performed separately for each dependency relation. A word pair is included only when embeddings are available for both words under the same relation, and a relation–subset combination is reported only if at least 50\% of the benchmark pairs are covered. Performance is measured using Spearman’s rank correlation between embedding-based cosine similarities and human judgements. For each correlation, we also report the number and proportion of matched pairs to make differences in vocabulary coverage explicit.

\begin{table*}[t]
\centering
\caption{Spearman's $\rho$ on the WordSim Similarity (Sim.) and
Relatedness (Rel.) subsets across dependency relations.
Bold values indicate the best-performing model for each
relation and subset.}
\label{tab:wordsim_correlations}
\small
\setlength{\tabcolsep}{5pt}
\begin{tabular}{lrrrrrrrr}
\hline
& \multicolumn{4}{c}{\textbf{Similarity}} 
& \multicolumn{4}{c}{\textbf{Relatedness}} \\
\cline{2-5}\cline{6-9}
\textbf{Relation} 
& \textbf{Rev.} & \textbf{Orig.} & \textbf{SVD} & \textbf{PPMI}
& \textbf{Rev.} & \textbf{Orig.} & \textbf{SVD} & \textbf{PPMI} \\
\hline
\texttt{chi\_compound} & .402 & .500 & .512 & \textbf{.565} & .304 & \textbf{.321} & .295 & .260 \\
\texttt{chi\_conj}     & .489 & .512 & .581 & \textbf{.602} & .206 & .241 & .285 & \textbf{.314} \\
\texttt{chi\_nmod}     & .422 & .445 & .452 & \textbf{.502} & .084 & .168 & .187 & \textbf{.204} \\
\texttt{chi\_nsubj}    & .489 & \textbf{.527} & .500 & .468 & .192 & \textbf{.285} & .222 & .249 \\
\texttt{chi\_obj}      & .502 & .497 & .551 & \textbf{.611} & .179 & .188 & .155 & \textbf{.251} \\
\texttt{chi\_obl}      & .445 & \textbf{.536} & .495 & .520 & .218 & .274 & \textbf{.312} & .252 \\
\texttt{pa\_amod}      & \textbf{.505} & .473 & .445 & .480 & .107 & .152 & .126 & \textbf{.224} \\
\texttt{pa\_case}      & .212 & .222 & .400 & \textbf{.478} & .007 & .121 & -.033 & \textbf{.149} \\
\texttt{pa\_cc}        & .085 & \textbf{.161} & .122 & .045 & \textbf{.007} & -.073 & .001 & -.052 \\
\texttt{pa\_compound}  & .387 & .454 & .494 & \textbf{.525} & .180 & .239 & \textbf{.380} & .278 \\
\texttt{pa\_conj}      & .502 & .522 & .540 & \textbf{.566} & \textbf{.324} & .311 & .271 & .296 \\
\texttt{pa\_det}       & .023 & -.082 & .175 & \textbf{.222} & -.018 & -.133 & -.045 & \textbf{.019} \\
\texttt{pa\_nmod}      & \textbf{.485} & .479 & .481 & .413 & .168 & .185 & \textbf{.293} & .244 \\
\texttt{pa\_punct}     & .107 & -.124 & .103 & \textbf{.160} & -.020 & -.174 & .032 & \textbf{.045} \\
\hline
\end{tabular}

\vspace{2pt}
\begin{minipage}{0.95\textwidth}
\footnotesize
\textit{Note:} Rev. = Reversed SGNS;
Orig. = Original-context SGNS;
SVD = PPMI-SVD (300 dimensions);
PPMI = PPMI-5000.
\end{minipage}
\end{table*}

After applying the 50\% coverage threshold, 14 dependency relations remain eligible for both subsets (Table~\ref{tab:wordsim_coverage}). Coverage ranges from 53.2\% to 87.3\%. For every reported relation--subset combination, the number of matched pairs is identical across the 4 semantic spaces. Consequently, differences in performance across models cannot be attributed to differences in benchmark coverage.

\begin{table}[htbp]
\centering
\caption{Coverage of the fourteen retained dependency
relations in the WordSim subsets. Counts are identical
across all 4 semantic spaces.}
\label{tab:wordsim_coverage}
\small
\setlength{\tabcolsep}{5pt}
\begin{tabular}{lcc}
\hline
\textbf{Relation} &
\textbf{Similarity} &
\textbf{Relatedness} \\
\hline
\texttt{chi\_compound} & 115/203 & 151/252 \\
\texttt{chi\_conj}     & 131/203 & 172/252 \\
\texttt{chi\_nmod}     & 133/203 & 188/252 \\
\texttt{chi\_nsubj}    & 112/203 & 148/252 \\
\texttt{chi\_obj}      & 133/203 & 178/252 \\
\texttt{chi\_obl}      & 139/203 & 192/252 \\
\texttt{pa\_amod}      & 145/203 & 195/252 \\
\texttt{pa\_case}      & 166/203 & 220/252 \\
\texttt{pa\_cc}        & 108/203 & 145/252 \\
\texttt{pa\_compound}  & 119/203 & 156/252 \\
\texttt{pa\_conj}      & 121/203 & 159/252 \\
\texttt{pa\_det}       & 151/203 & 197/252 \\
\texttt{pa\_nmod}      & 120/203 & 160/252 \\
\texttt{pa\_punct}     & 159/203 & 203/252 \\
\hline
\end{tabular}
\end{table}

\paragraph{Cross-Method Representational Similarity Analysis (RSA)}
\label{sec:appendix_rsa}
For each dependency relation, we select the 200 most frequent dependency-conditioned items shared across all 4 representations, using their relation-specific frequencies. Relations containing fewer than 200 shared items are excluded, resulting in 50 dependency relations. For each model $m$ and dependency relation $r$, we construct a pairwise cosine-similarity matrix $S^{(m,r)}$ over the 200 shared items. For each pair of representations $(m_1,m_2)$, RSA is calculated as the Spearman rank correlation between their corresponding similarity values:

\begin{equation}
    \scalebox{0.85}{$\displaystyle
    \mathrm{RSA}_{r}(m_1,m_2)
    =
    \rho_s
    \left(
        \operatorname{vec}_{\triangle}(S^{(m_1,r)}),
        \operatorname{vec}_{\triangle}(S^{(m_2,r)})
    \right)
    $}
\end{equation}
where $\operatorname{vec}_{\triangle}$ extracts the upper-triangular elements of a matrix.

Higher RSA values indicate greater agreement between models in the relative ordering of lexical similarities. 6 model-pair comparisons are performed for each of the 50 relations, yielding 300 relation-level RSA scores. These scores are subsequently summarised using their macro-mean, median, and standard deviation across relations.

\begin{table}[htbp]
\centering
\small
\setlength{\tabcolsep}{3pt}
\caption{Cross-method representational similarity across 50 dependency relations ($N=200$). Values are macro-level descriptive statistics of relation-specific Spearman correlations.}
\label{tab:cross_method_rsa}
\begin{tabular}{lccc}
\hline
\textbf{Model comparison} &
\textbf{Mean $\rho$} &
\textbf{Median $\rho$} &
\textbf{SD} \\
\hline
PPMI-SVD / PPMI-5000
    & .791 & .802 & .122 \\
Rev. / Orig. SGNS
    & .774 & .820 & .140 \\
\hline
Orig. SGNS / PPMI-SVD
    & .654 & .721 & .245 \\
Rev. SGNS / PPMI-SVD
    & .610 & .682 & .244 \\
Orig. SGNS / PPMI-5000
    & .613 & .702 & .267 \\
Rev. SGNS / PPMI-5000
    & .581 & .668 & .260 \\
\hline
\end{tabular}
\end{table}

\begin{table}[t]
\centering
\small
\setlength{\tabcolsep}{4pt}
\caption{Dependency relations with the highest and lowest cross-family representational agreement. Values are mean Spearman correlations across 4 SGNS--PPMI comparisons.}
\label{tab:relation_specific_rsa}
\begin{tabular}{lclc}
\hline
\multicolumn{2}{c}{\textbf{Highest agreement}} &
\multicolumn{2}{c}{\textbf{Lowest agreement}} \\
\hline
\textbf{Relation} & $\boldsymbol{\rho}$ &
\textbf{Relation} & $\boldsymbol{\rho}$ \\
\hline
\texttt{pa\_conj}    & .911 & \texttt{pa\_det}  & -.029 \\
\texttt{chi\_conj}   & .909 & \texttt{pa\_mark} & .058 \\
\texttt{chi\_nummod} & .890 & \texttt{pa\_cc}   & .125 \\
\texttt{chi\_nmod}   & .856 & \texttt{pa\_aux}  & .126 \\
\texttt{pa\_xcomp}   & .835 & \texttt{pa\_cop}  & .135 \\
\hline
\end{tabular}
\end{table}

\newpage
\subsection{Linguistic Interpretation}

\begin{figure*}[htbp]
    \centering
    \includegraphics[
        width=\textwidth,
        height=0.9\textheight,
        keepaspectratio
    ]{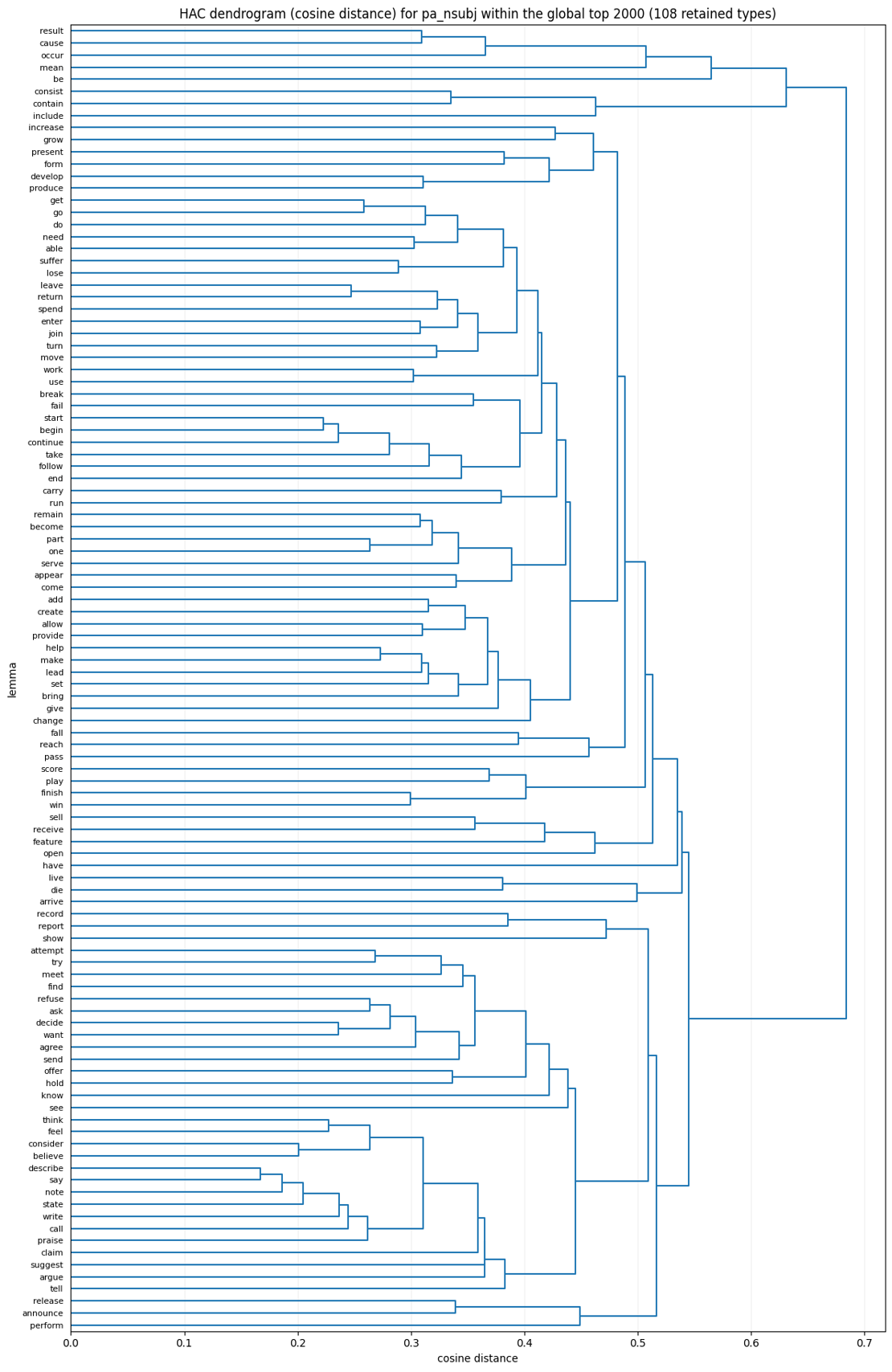}
    \caption{HAC dendrogram for \texttt{pa\_nsubj} of Original-context SGNS}
    \label{fig:pa_nsubj_dendrogram}
\end{figure*}

\begin{table}[htbp]
\centering
\small
\setlength{\tabcolsep}{4pt}
\begin{tabularx}{\columnwidth}{
    @{}ll>{\raggedright\arraybackslash}X@{}
}
\toprule
\textbf{Type} & \textbf{POS} & \textbf{Lemmas} \\
\midrule
Within-POS & Noun &
bank, paper, channel, market, branch \\
& Verb &
hold, charge, turn, see, find \\
& Adjective &
hard, heavy, certain, late, responsible \\
\midrule
Cross-POS & Noun &
file, match, face, head, chair \\
& Verb &
record, plant, drive, call, watch \\
& Adjective &
fine, right, present, mean, last \\
\bottomrule
\end{tabularx}
\caption{30 selected lemmas for qualitative analysis.}
\label{tab:polysemy_candidates}
\end{table}

The following tables report 2 selected dependency relations for each of 30 lemmas. For each relation, the 5 nearest neighbours from each of the 4 models are reported in decreasing cosine-similarity order. Reading labels are qualitative interpretations rather than independently annotated gold senses; cases labelled \emph{weak} do not show a clear semantic contrast in the selected relations.

% \paragraph{Nearest-neighbour exploration}

\clearpage
\begin{table*}[t]
\centering
\caption{Within-POS nouns examples. 5 neighbours are shown in decreasing cosine-similarity order.}
\label{tab:full_within_nouns}
\scriptsize
\setlength{\tabcolsep}{3pt}
\renewcommand{\arraystretch}{1.15}
\begin{tabular}{p{1.8cm} l p{2.65cm} p{2.65cm} p{2.65cm} p{2.65cm}}
\hline
\textbf{Reading} & \textbf{Relation} & \textbf{Reversed SGNS} & \textbf{Original-context SGNS} & \textbf{PPMI-SVD} & \textbf{PPMI-5000} \\
\hline
bank: financial &
\texttt{chi\_compound} &
insurance, mortgage, booking, government, bankruptcy &
mortgage, gate, booking, wire, KGB &
government, service, property, estate, council &
government, banking, service, company, Bank \\
\hline
bank: river edge &
\texttt{chi\_obl} &
tributary, estuary, foothill, valley, ravine &
valley, estuary, Plain, Firth, shore &
hill, river, shore, valley, River &
River, town, area, shore, south \\
\hline
paper: material &
\texttt{chi\_compound} &
flour, cardboard, coin, textile, chocolate &
cloth, flour, deposit, wool, cardboard &
cardboard, rubber, print, plastic, pottery &
cloth, silk, plastic, textile, flour \\
\hline
paper: publication &
\texttt{chi\_appos} &
Times, publication, News, Review, Journal &
Times, Journal, publication, Review, journal &
newspaper, publication, Times, Review, magazine &
newspaper, pp., Times, News, Sun \\
\hline
channel: broadcasting &
\texttt{chi\_appos} &
NBC, ABC, FM, f.m., Radio &
FM, f.m., NBC, ABC, TV &
network, TV, ABC, f.m., Radio &
TV, ABC, NBC, affiliate, network \\
\hline
channel: passage &
\texttt{chi\_obj} &
corridor, Channel, pipeline, log, roadway &
pipeline, sand, frontier, waterway, runway &
trail, corridor, field, lane, canal &
road, route, river, highway, line \\
\hline
market: economic sector &
\texttt{chi\_compound} &
tourism, estate, sporting, commodity, lifestyle &
mortgage, sector, resource, industry, tourism &
industry, business, consumer, tourism, product &
industry, retail, consumer, business, sale \\
\hline
market: trade/distribution &
\texttt{pa\_nmod} &
distributor, export, retailer, sell, utility &
price, sale, manufacturer, export, goods &
goods, manufacturer, seller, store, retailer &
trade, export, sale, city, port \\
\hline
branch: botanical &
\texttt{chi\_conj} &
line, trunk, shrub, stem, section &
trunk, line, shrub, stem, basin &
line, route, Bridge, road, junction &
Creek, trunk, root, line, Road \\
\hline
branch: corporate &
\texttt{pa\_nmod:poss} &
stock, merger, monopoly, subsidiary, franchise &
subsidiary, stock, founder, network, franchise &
archive, board, chairman, merger, asset &
program, headquarters, founder, department, subsidiary \\
\hline
\end{tabular}
\end{table*}

\begin{table*}[t]
\centering
\caption{Within-POS verbs examples. 5 neighbours are shown in decreasing cosine-similarity order.}
\label{tab:full_within_verbs}
\scriptsize
\setlength{\tabcolsep}{3pt}
\renewcommand{\arraystretch}{1.15}
\begin{tabular}{p{1.8cm} l p{2.65cm} p{2.65cm} p{2.65cm} p{2.65cm}}
\hline
\textbf{Reading} & \textbf{Relation} & \textbf{Reversed SGNS} & \textbf{Original-context SGNS} & \textbf{PPMI-SVD} & \textbf{PPMI-5000} \\
\hline
hold: host event &
\texttt{chi\_acl} &
take, contest, host, preside, benefit &
organised, contest, host, preside, take &
host, organised, attend, organize, celebrate &
take, host, attend, organised, call \\
\hline
hold: stance &
\texttt{pa\_ccomp} &
argue, establish, conclude, maintain, believe &
argue, rule, maintain, conclude, assert &
rule, contend, maintain, determine, establish &
argue, believe, suggest, claim, state \\
\hline
charge: accusation &
\texttt{chi\_obj} &
complaint, rumor, allegation, ultimatum, objection &
barrage, torpedo, warhead, allegation, complaint &
case, call, complaint, order, demand &
claim, allegation, accusation, attack, case \\
\hline
charge: financial &
\texttt{pa\_obj} &
fine, bribe, pay, station, alert &
dislodge, unload, bribe, outrun, slaughter &
pay, count, cost, fine, allocate &
pay, attack, cost, worth, collect \\
\hline
turn: sequence &
\texttt{chi\_obl} &
stage, example, period, time, day &
way, what, whole, process, relocation &
process, consequence, course, extent, effort &
which, time, result, end, period \\
\hline
turn: direction &
\texttt{pa\_advmod} &
shift, accelerate, bend, angle, meander &
shift, bend, angle, tilt, swing &
push, move, extend, travel, shift &
pass, move, lead, run, make \\
\hline
see: propositional understanding &
\texttt{pa\_ccomp} &
know, claim, believe, realize, add &
think, tell, add, know, find &
know, explain, sure, report, find &
find, state, say, tell, believe \\
\hline
see: perception &
\texttt{pa\_obl:agent} &
believe, view, watch, notice, think &
view, watch, believe, think, notice &
view, regard, believe, admire, interprete &
view, watch, regard, receive, consider \\
\hline
find: discovery &
\texttt{pa\_nmod} &
artefact, woodland, cache, artifact, topography &
excavation, artifact, dating, locality, dinosaur &
artefact, artifact, fragment, burial, excavation &
excavation, discovery, remains, site, fossil \\
\hline
find: conclusion &
\texttt{pa\_ccomp} &
conclude, show, believe, suggest, state &
believe, prove, suggest, show, claim &
show, suggest, claim, mean, conclude &
show, believe, suggest, state, note \\
\hline
\end{tabular}
\end{table*}

\begin{table*}[t]
\centering
\caption{Within-POS adjectives examples. 5 neighbours are shown in decreasing cosine-similarity order.}
\label{tab:full_within_adjectives}
\scriptsize
\setlength{\tabcolsep}{3pt}
\renewcommand{\arraystretch}{1.15}
\begin{tabular}{p{1.8cm} l p{2.65cm} p{2.65cm} p{2.65cm} p{2.65cm}}
\hline
\textbf{Reading} & \textbf{Relation} & \textbf{Reversed SGNS} & \textbf{Original-context SGNS} & \textbf{PPMI-SVD} & \textbf{PPMI-5000} \\
\hline
hard: physical &
\texttt{chi\_amod} &
soft, process, tougher, shiny, tough &
tough, tougher, slippery, tinged, flavor &
soft, solid, melt, loose, coarse &
soft, tough, solid, more, color \\
\hline
hard: difficulty &
\texttt{pa\_ccomp} &
difficult, easy, impossible, affect, hope &
easy, impossible, difficult, discuss, affect &
difficult, impossible, easy, help, teach &
difficult, easy, impossible, state, say \\
\hline
heavy: weight &
\texttt{chi\_amod} &
light, massed, torrential, cement, artillery &
light, massed, torrential, cement, excess &
light, explosive, less, minimal, more &
light, more, additional, less, massive \\
\hline
heavy: intensity/mixed &
\texttt{pa\_obl} &
strong, severe, minor, cause, light &
light, minor, strong, wet, cause &
strong, severe, light, low, minimal &
cause, high, occur, produce, large \\
\hline
certain: specificity &
\texttt{chi\_amod} &
various, such, many, different, same &
such, various, specific, particular, specify &
particular, various, same, different, such &
specific, various, different, such, many \\
\hline
certain: epistemic &
\texttt{pa\_ccomp} &
unlikely, question, claim, conclude, realize &
allege, question, realize, observe, clear &
unlikely, likely, appear, uncertain, presume &
know, believe, suggest, claim, confirm \\
\hline
late: temporal (weak) &
\texttt{chi\_amod} &
early, previous, mid, recent, subsequent &
early, mid, recent, subsequent, contemporary &
early, mid, mid-19th, mid-18th, mid-20th &
early, subsequent, recent, previous, mid \\
\hline
responsible: causation/mixed (weak) &
\texttt{pa\_obl} &
contribute, involved, provide, result, require &
need, result, contribute, aid, essential &
assist, contribute, aid, result, need &
result, contribute, lead, involved, assist \\
\hline
\end{tabular}
\end{table*}

\begin{table*}[t]
\centering
\caption{Cross-POS noun-centred examples. 5 neighbours are shown in decreasing cosine-similarity order.}
\label{tab:full_cross_nouns}
\scriptsize
\setlength{\tabcolsep}{3pt}
\renewcommand{\arraystretch}{1.15}
\begin{tabular}{p{1.8cm} l p{2.65cm} p{2.65cm} p{2.65cm} p{2.65cm}}
\hline
\textbf{Reading} & \textbf{Relation} & \textbf{Reversed SGNS} & \textbf{Original-context SGNS} & \textbf{PPMI-SVD} & \textbf{PPMI-5000} \\
\hline
file: document &
\texttt{chi\_obj} &
e-mail, diary, inventory, email, Spirit &
diary, gem, software, poison, Spirit &
sample, fragment, photograph, tape, photo &
document, data, information, map, tape \\
\hline
file: submit or lodge &
\texttt{pa\_obj} &
lodge, print, forward, invalidate, submit &
lodge, forward, retract, circulate, rescind &
issue, forward, submit, retract, appeal &
lodge, submit, withdraw, dismiss, issue \\
\hline
match: sporting noun &
\texttt{chi\_nmod} &
bout, game, fixture, final, tournament &
game, fixture, final, contest, bout &
round, final, win, tournament, contest &
game, final, tournament, event, season \\
\hline
match: correspond &
\texttt{pa\_obj} &
describe, duplicate, sum, vary, fitting &
fit, project, duplicate, discern, fitting &
change, alter, adjust, fit, display &
have, change, display, reflect, use \\
\hline
face: body part &
\texttt{chi\_obj} &
breast, sand, fence, cylinder, trunk &
breast, throat, paint, rock, sand &
eye, hair, head, breast, tail &
eye, head, nose, hand, image \\
\hline
face: confront &
\texttt{pa\_obj} &
overcome, fear, eliminate, avoid, over &
over, defeat, eliminate, confront, overcome &
eliminate, defeat, beat, pit, challenge &
defeat, beat, avoid, eliminate, overcome \\
\hline
head: noun &
\texttt{chi\_compound} &
neck, assistant, thigh, toe, shoulder &
thigh, neck, wrist, rib, assistant &
shoulder, nose, tail, neck, face &
assistant, back, knee, shoulder, body \\
\hline
head: lead &
\texttt{pa\_obj} &
chair, reorganize, curve, convene, center &
reorganise, chair, reorganize, curve, center &
chair, reorganize, restructure, reorganise, disband &
chair, join, lead, organize, establish \\
\hline
chair: furniture &
\texttt{chi\_conj} &
table, treasurer, bench, lamp, ladder &
treasurer, bench, lamp, ladder, table &
chairman, president, board, treasurer, table &
president, professor, chairman, bench, leader \\
\hline
chair: preside &
\texttt{chi\_acl} &
preside, oversee, coordinate, prosecute, authorize &
preside, oversee, appoint, coordinate, organise &
preside, oversee, head, investigate, charge &
head, task, appoint, preside, establish \\
\hline
\end{tabular}
\end{table*}

\begin{table*}[t]
\centering
\caption{Cross-POS verb-centred examples. 5 neighbours are shown in decreasing cosine-similarity order.}
\label{tab:full_cross_verbs}
\scriptsize
\setlength{\tabcolsep}{3pt}
\renewcommand{\arraystretch}{1.15}
\begin{tabular}{p{1.8cm} l p{2.65cm} p{2.65cm} p{2.65cm} p{2.65cm}}
\hline
\textbf{Reading} & \textbf{Relation} & \textbf{Reversed SGNS} & \textbf{Original-context SGNS} & \textbf{PPMI-SVD} & \textbf{PPMI-5000} \\
\hline
record: noun &
\texttt{chi\_nmod} &
time, album, work, season, music &
album, time, label, work, recording &
recording, label, album, title, work &
album, work, book, song, recording \\
\hline
record: verb &
\texttt{pa\_obj} &
recording, remix, perform, co-produce, preview &
recording, produce, perform, compare, co-produce &
recording, sang, perform, sing, si &
recording, post, perform, report, produce \\
\hline
plant: organism &
\texttt{chi\_conj} &
shrub, invertebrate, grasses, herb, fungi &
invertebrate, grasses, shrub, fungi, vegetable &
grasses, grass, tree, crop, flower &
fungi, insect, tree, fruit, fish \\
\hline
plant: action &
\texttt{pa\_obj} &
uproot, detonate, store, harvest, spray &
harvest, smuggle, uproot, detonate, fabricate &
harvest, grow, chew, uproot, eat &
attach, store, detonate, destroy, burn \\
\hline
drive: machinery &
\texttt{chi\_compound} &
V8, horsepower, handheld, mw, plug &
horsepower, plug, mw, V8, axle &
gear, launch, plug, lift, turbine &
launch, propulsion, horsepower, engine, diesel \\
\hline
drive: propel &
\texttt{pa\_obj} &
push, propel, dislodge, ambush, choke &
propel, keep, stop, push, control &
push, force, ambush, keep, trap &
force, push, stop, attack, keep \\
\hline
call: statement &
\texttt{chi\_obj} &
assertion, endorsement, signal, denial, plea &
message, edict, complaint, assertion, denial &
complaint, challenge, warning, suggestion, comment &
complaint, criticism, argument, request, signal \\
\hline
call: label &
\texttt{pa\_xcomp} &
dub, term, label, consider, name &
dub, term, label, brand, name &
label, term, brand, consider, dub &
consider, name, deem, title, find \\
\hline
watch: alert &
\texttt{pa\_compound} &
first, alert, advisory, forecasting, nine &
warning, Seven, auction, Beta, Boot &
warning, alert, relief, advisory, week &
warning, advisory, alert, forecasting, intensity \\
\hline
watch: observe &
\texttt{pa\_obj} &
see, tape, film, cheer, view &
film, see, stop, cheer, filming &
filming, film, direct, air, broadcast &
see, film, air, direct, call \\
\hline
\end{tabular}
\end{table*}

\begin{table*}[t]
\centering
\caption{Cross-POS adjective-centred examples. 5 neighbours are shown in decreasing cosine-similarity order.}
\label{tab:full_cross_adjectives}
\scriptsize
\setlength{\tabcolsep}{3pt}
\renewcommand{\arraystretch}{1.15}
\begin{tabular}{p{1.8cm} l p{2.65cm} p{2.65cm} p{2.65cm} p{2.65cm}}
\hline
\textbf{Reading} & \textbf{Relation} & \textbf{Reversed SGNS} & \textbf{Original-context SGNS} & \textbf{PPMI-SVD} & \textbf{PPMI-5000} \\
\hline
fine: quality &
\texttt{chi\_amod} &
quality, coarse, superb, exquisite, elegant &
exquisite, elegant, quality, pleasing, colourful &
polished, brilliant, elegant, soft, stain &
good, famous, excellent, brilliant, beautiful \\
\hline
fine: penalty &
\texttt{chi\_obj} &
dividend, subsidy, Assent, penalty, tax &
dividend, Assent, condemnation, bribe, pension &
tax, fee, sanction, suspension, writ &
penalty, tax, ban, fee, sanction \\
\hline
right: adjective &
\texttt{chi\_amod} &
left, sore, leave, medial, wrong &
left, leave, oblique, wrong, sever &
left, broken, sore, muscular, bruise &
left, leave, wrong, front, different \\
\hline
right: entitlement &
\texttt{chi\_obj} &
copyright, liberty, privilege, dignity, citizenship &
privilege, dignity, liberty, patent, copyright &
sovereignty, ownership, control, privilege, citizenship &
privilege, sovereignty, independence, freedom, autonomy \\
\hline
present: current &
\texttt{chi\_amod} &
current, new, different, same, pre-exist &
current, serve, convenient, pre-exist, new &
current, certain, typical, same, particular &
current, new, different, certain, exist \\
\hline
present: verbal &
\texttt{pa\_obj} &
see, compare, find, tout, view &
see, compare, address, find, accept &
view, consider, compare, base, characterize &
consider, describe, introduce, show, discuss \\
\hline
mean: adjective &
\texttt{chi\_amod} &
blonde, average, superhuman, gentle, stereotypical &
average, freezing, virtuous, gentle, unnatural &
vary, measure, approximate, comparable, average &
average, maximum, minimum, approximate, vary \\
\hline
mean: verb &
\texttt{pa\_ccomp} &
write, say, find, suggest, believe &
show, find, determine, indicate, suggest &
show, indicate, find, ensure, determine &
show, suggest, believe, state, find \\
\hline
last: final &
\texttt{chi\_amod} &
first, second, previous, next, final &
next, final, previous, only, first &
first, next, second, third, final &
first, next, final, second, previous \\
\hline
last: endure &
\texttt{pa\_nsubj} &
end, continue, begin, follow, prove &
end, begin, ensue, continue, commence &
over, culminate, ensue, commence, progress &
end, result, begin, over, occur \\
\hline
\end{tabular}
\end{table*}

\end{document}